\documentclass[10pt,twocolumn,letterpaper]{article}

\usepackage{iccv}
\usepackage{times}
\usepackage{epsfig}
\usepackage{graphicx}
\usepackage{amsmath}
\usepackage{amssymb}
\usepackage{subfigure}
\usepackage{color}
\usepackage{booktabs}
\usepackage{multirow}
\usepackage{makecell}

\renewcommand{\thefootnote}{}

\newcommand{\dz}[1]{\textcolor[rgb]{0,0,0}{#1}}
\newcommand{\cm}[1]{\textcolor[rgb]{0,0,0}{#1}}
\newcommand{\cmx}[1]{\textcolor[rgb]{0, 0, 0}{#1}}

\usepackage[pagebackref=true,breaklinks=true,letterpaper=true,colorlinks,bookmarks=false, urlcolor=black]{hyperref}

\iccvfinalcopy 

\ificcvfinal\fi

\begin{document}

\title{Location-aware Single Image Reflection Removal}


\author{
Zheng Dong$^{1}$\ \ \ Ke Xu$^{2,4}$\ \ \ Yin Yang$^{3}$\ \ \ Hujun Bao$^{1}$\ \ \ Weiwei Xu$^{*1}$\ \ \ Rynson W.H. Lau$^{4}$
\vspace{3pt} \\
\normalsize $^{1}$State Key Lab of CAD\&CG, Zhejiang University \quad \quad
$^{2}$Shanghai Jiao Tong University \\
\normalsize $^{3}$Clemson University \quad \quad
$^{4}$City University of Hong Kong \\
\small zhengdong@zju.edu.cn,\ \  kkangwing@gmail.com,\ \ yin5@clemson.edu,\ \ \{bao, xww\}@cad.zju.edu.cn,\ \ Rynson.Lau@cityu.edu.hk
\vspace{-10pt}
}

\maketitle
\ificcvfinal\thispagestyle{empty}\fi

\begin{abstract}
\cm{This paper proposes a novel location-aware deep-learning-based single image reflection removal method. Our network has a reflection detection module to regress a probabilistic reflection confidence map, taking multi-scale Laplacian features as inputs. This probabilistic map tells if a region is reflection-dominated or transmission-dominated, and it is used as a cue for the network to control the feature flow when predicting the reflection and transmission layers. We design our network as a recurrent network to progressively refine reflection removal results at each iteration. The novelty is that we leverage Laplacian kernel parameters to emphasize the boundaries of strong reflections. It is beneficial to strong reflection detection and substantially improves the quality of reflection removal results. Extensive experiments verify the superior performance of the proposed method over state-of-the-art approaches. Our code and the pre-trained model can be found at \href{https://github.com/zdlarr/Location-aware-SIRR}{ \textcolor{blue}{https://github.com/zdlarr/Location-aware-SIRR}}}.
\end{abstract}
\footnote{$^{*}$Corresponding author. }
\vspace{-2pt}
\section{Introduction}
\label{sec:introduction}

Reflections often occur when images are photographed through reflective and transparent media (\eg, glass).
Removing undesired reflections enhances the image quality and benefits many follow-up computer vision tasks, such as image classification. In reflection removal, an image $\mathbf{I}$ with reflections can be modeled as the weighted additive composition of a transmission layer $\mathbf{T}$ and a reflection layer $\mathbf{R}$. Precisely, following the alpha blending model in~\cite{fan2017generic, li2020single, zhang2018single}, we express the composition procedure as:
{
\setlength\abovedisplayskip{3pt}
\setlength\belowdisplayskip{3pt}
\begin{equation}
\mathbf{I} = \mathbf{W} \circ \mathbf{T} +  \mathbf{R},
\label{eq:reflection_decomposition}
\end{equation}
where $\mathbf{W}$ here is an alpha blending mask and $\circ$ indicates the element-wise multiplication. This model is designed to approximate the complex physical mechanisms involved in forming images with reflections.}
\begin{figure}[t]
\setlength{\tabcolsep}{.8pt}
 \centering
    \begin{tabular}{ccc}
    \vspace{-5pt}
    \includegraphics[width=0.325\linewidth,keepaspectratio]{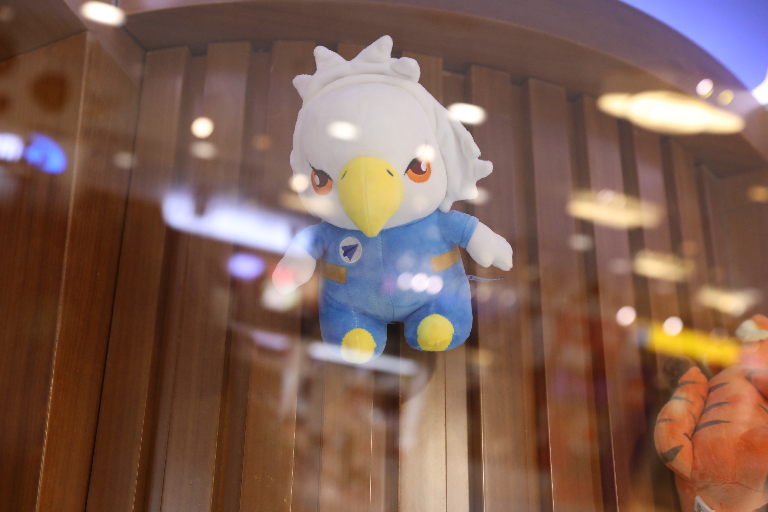}&
    \includegraphics[width=0.325\linewidth,keepaspectratio]{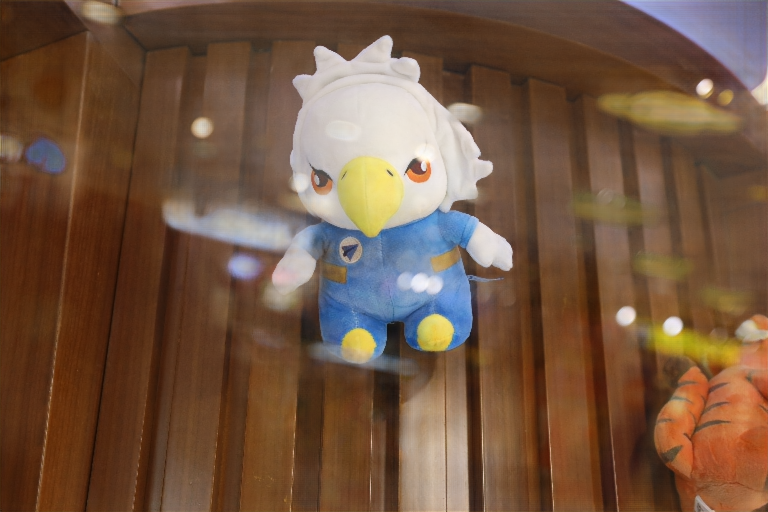}&
    \includegraphics[width=0.325\linewidth,keepaspectratio]{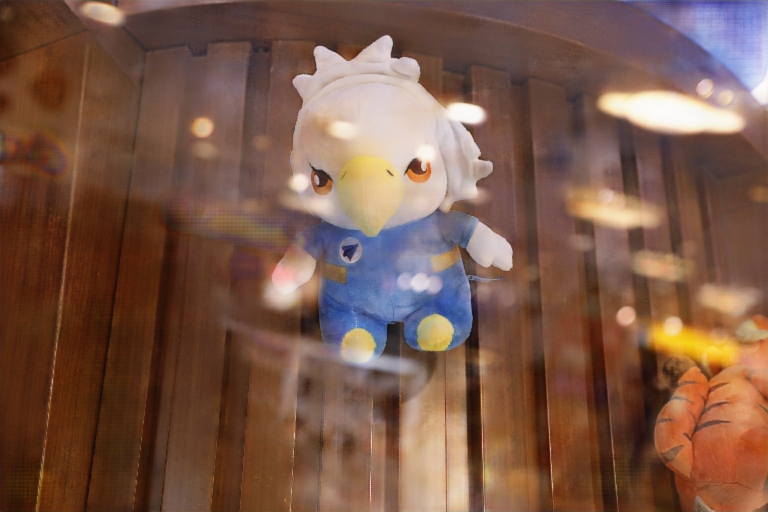} \\
    \scriptsize (a) Input & 
    \scriptsize (b) Zhang~\etal~\cite{zhang2018single} & 
    \scriptsize (c) RMNet~\cite{wen2019single} \\
    
    \vspace{-5pt}
    \includegraphics[width=0.325\linewidth,keepaspectratio]{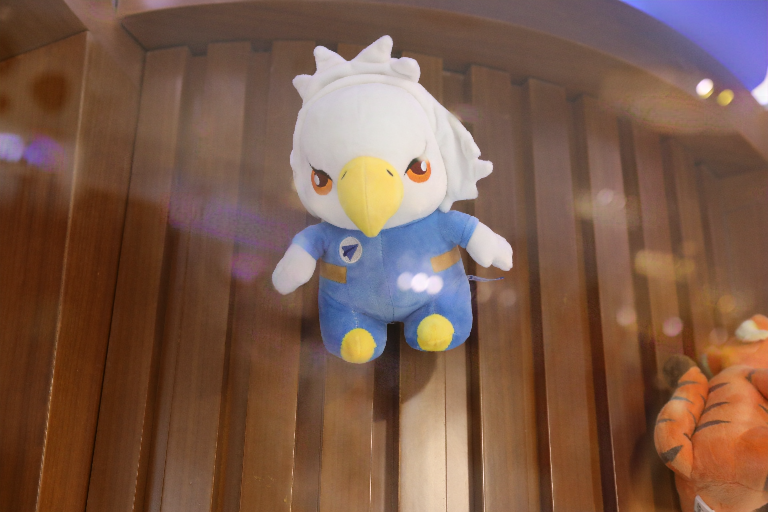} &
    \includegraphics[width=0.325\linewidth,keepaspectratio]{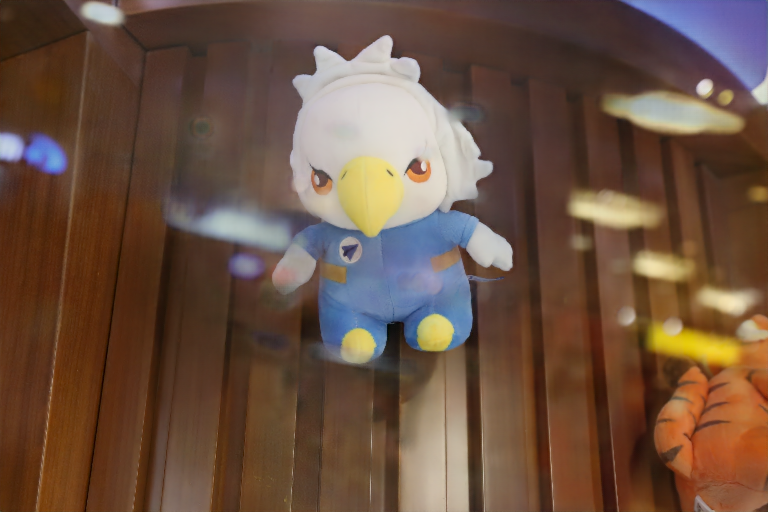} &
    \includegraphics[width=0.325\linewidth,keepaspectratio]{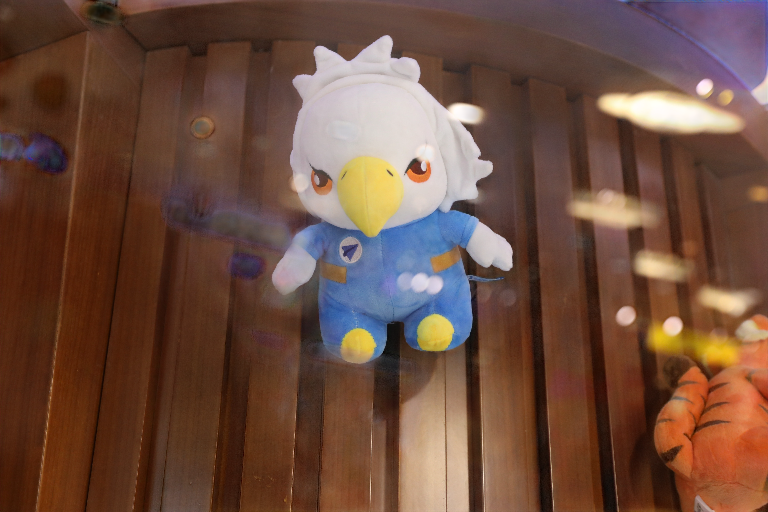} \\
      \scriptsize (d) ERRNet~\cite{wei2019single} &
      \scriptsize (e) CoRRN~\cite{wan2019corrn} & 
      \scriptsize (f) IBCLN~\cite{li2020single}  \\
    
    \vspace{-5pt}
    \includegraphics[width=0.325\linewidth,keepaspectratio]{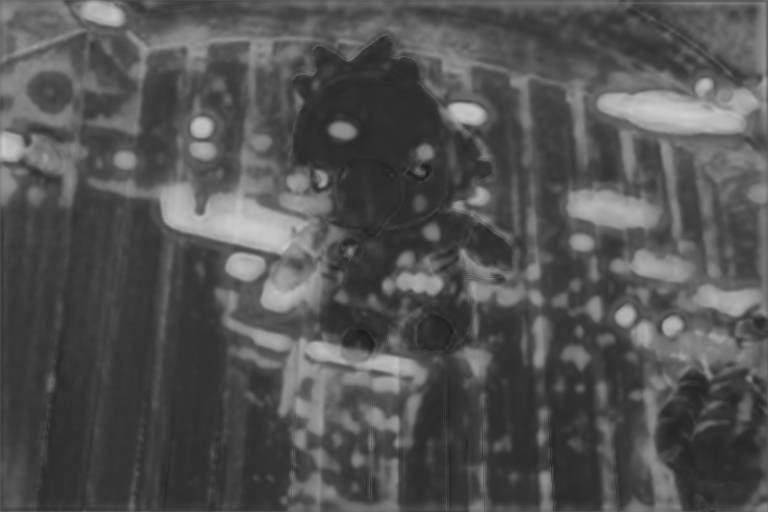}&
    \includegraphics[width=0.325\linewidth,keepaspectratio]{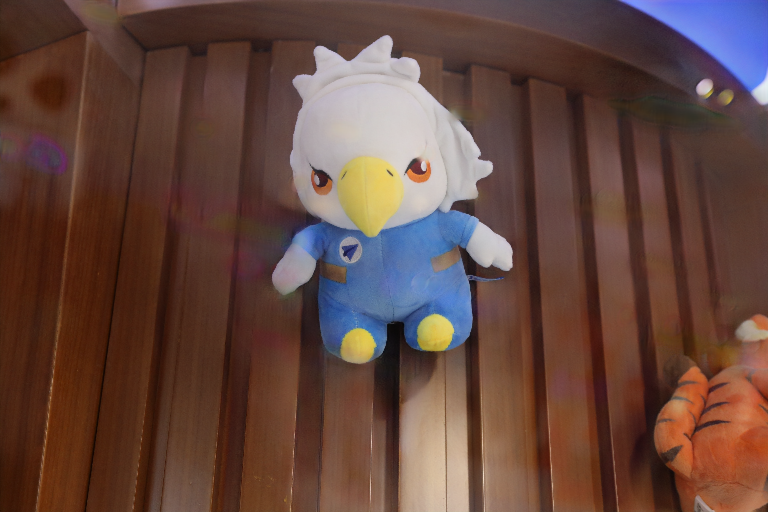}&
    \includegraphics[width=0.325\linewidth,keepaspectratio]{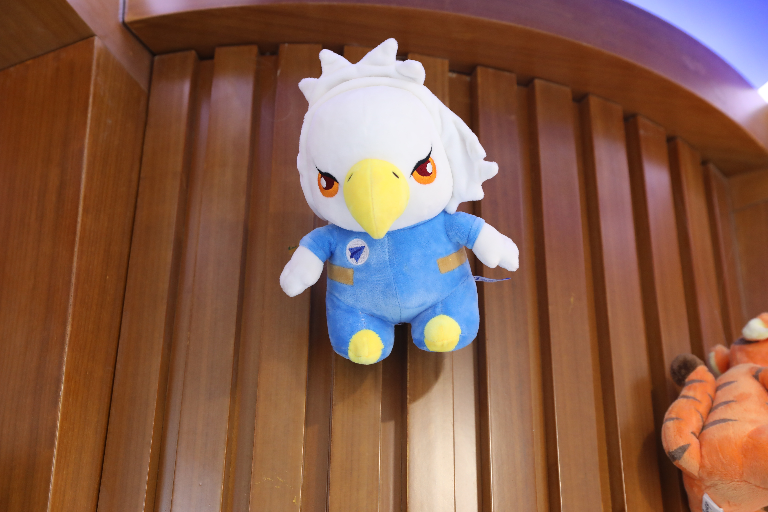} \\
      \scriptsize (g) RCMap & 
      \scriptsize (h) Ours  & 
      \scriptsize (i) Ground truth  \\
       
    \end{tabular}
    \vspace{1pt}
    \caption{State-of-the-art methods (b,c,d,e,f) typically fail to recover high-quality transmission layers from strong reflections, \eg, the highlights. Our method addresses this problem by learning the reflection confidence map\ (RCMap) for the detection of the reflection-dominated regions\ (g) and reflection removal\ (h). Images\ (a) and (i) are from~\cite{wei2019single}.}
\label{fig:teaser}
\vspace{-16pt}
\end{figure}


The task of single image reflection removal (SIRR) is to recover $\mathbf{T}$ from a given image $\mathbf{I}$. It is an ill-posed problem since the number of unknowns is much more than the number of equations derived from Eq.~\ref{eq:reflection_decomposition}.
Therefore, priors are necessary to constrain the solution space, such as natural image gradient sparsity~\cite{Levin2007,levin2003learning}, ghosting cues for thick glasses~\cite{shih2015reflection}, and relative smoothness that assumes the refection layer is smoother than the transmission layer~\cite{li2014single,yang2019fast}. To disambiguate the restoration of $\mathbf{T}$ and $\mathbf{R}$ in the gradient domain, several works propose first to determine the locations of reflection-dominated and transmission-dominated pixels, and then exploit different constraints at different locations to improve the reflection removal results~\cite{Levin2007,wan2018region, wan2016depth}. While these methods are sensitive to the selection of hyperparameters, for instance, the commonly used gradient magnitude threshold, the detected location information is proven to be useful to handle strong reflections. 
\cmx{Therefore, such location information is expected to be beneficial for the neural networks to learn how the reflection information is encoded in the features. However, it is rarely investigated in deep learning-based SIRR methods~\cite{fan2017generic, jin2018learning, kim2020single, li2020single, wei2019single, wen2019single, yang2018seeing, zhang2018single}}.
This might cause ambiguities when strong reflections appear. We observe that state-of-the-art SIRR methods typically fail to recover high-quality transmission layers in such cases.
\cm{This paper proposes a location-aware deep learning-based SIRR method for generic reflection removal. Our network incorporates a novel reflection detection module (RDM) to detect reflection-dominated regions via learning multi-scale Laplacian features. The output of RDM is a probabilistic reflection confidence map (RCMap), which controls the subsequent feature flow, resulting in significantly improved SIRR results.
\cmx{Although this detection-and-removal strategy has been explored in shadow analysis~\cite{ding2019argan, hu2019direction, hu2019mask, khan2015automatic, qu2017deshadownet, wang2018stacked} and rain removal~\cite{garg2004detection,kim2013single,qian2018attentive,yang2019joint,yasarla2019uncertainty,you2015adherent} tasks, they mainly exploit features learned in the RGB domain, which are different from our approach in two aspects.} 
First, motivated by the relative smoothness prior in~\cite{li2014single}, which assumes that the characteristics of $\mathbf{T}$ and $\mathbf{R}$ in the gradient domain are different, we use the learned Laplacian kernel to emphasize the boundaries of strong reflections and suppress low-frequency reflections, which is beneficial to improve the quality of RCMaps. Second, RDM is trained without ground-truth reflection-dominated region masks. This can avoid the difficulty of defining or labeling reflection-dominated regions. The inverse RCMap, \ie, 1 - RCMap, can serve as the weight map $\mathbf{W}$ in Eq.~\ref{eq:reflection_decomposition} and be used to indicate the transmission-dominated regions. It motivates us to use Eq.~\ref{eq:reflection_decomposition} as a loss to control the RDM training.}

\cm{
Our proposed network iteratively restores the transmission layer from the corrupted input. In each iteration, we formulate the restoration process in a removal-by-detection manner. It first detects the reflection-dominated regions based on the RCMap and then predicts the whole reflection layer by suppressing the transmission information. Afterwards, it restores the transmission layer by jointly leveraging the transmission-dominant regions and the predicted reflection layer. Such a network design decomposes the complicated SIRR problem into sub-problems and considers the mutual dependence between reflection and transmission.
\cmx{As illustrated in Fig.~\ref{fig:teaser}, our network can effectively remove strong reflected highlights.}
}

In summary, the main contributions of our work are:
\vspace{-4pt}
\begin{itemize}
\setlength\itemsep{-0.15em}
    \item We propose a novel SIRR method that iteratively restores the transmission layer from the corrupted input image. At each iteration, the restoration process is formulated in a removal-by-detection sequential manner.
    \item We propose a novel reflection detection module (RDM) to locate the reflection-dominated regions. It learns a group of multi-scale Laplacian kernel parameters to exploit reflection boundary information.
    \item Extensive experiments show that the proposed location-aware neural network outperforms state-of-the-art SIRR methods in removing strong reflections.
\end{itemize}

\begin{figure*}[t]
\centering
\includegraphics[width=\linewidth]{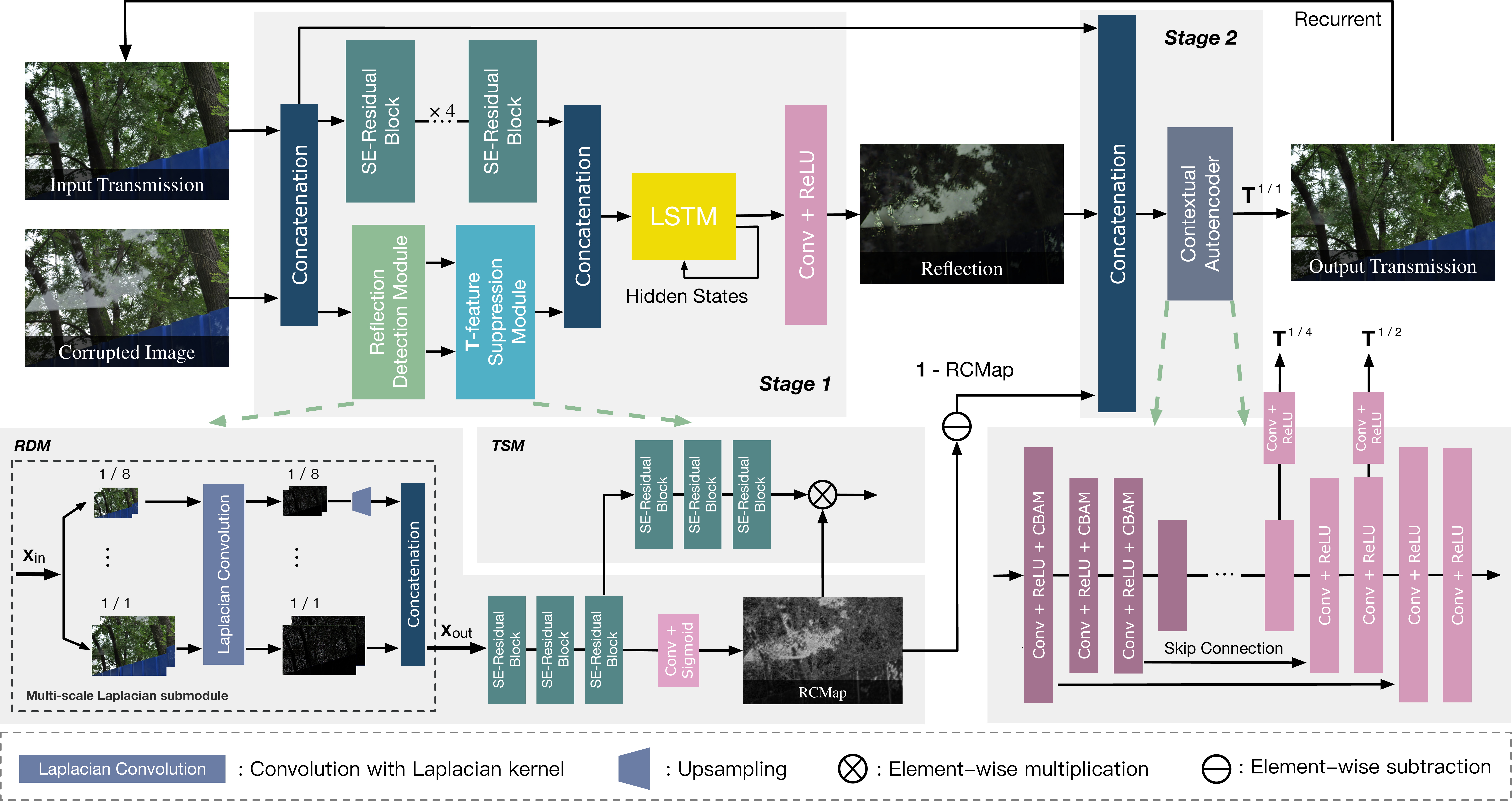}
\caption{The architecture of our recurrent SIRR network. Stage 1: predict the RCMap and the reflection layer. ``x4'' indicates that the SE (Squeeze-and-Excitation) residual block~\cite{hu2018squeeze} is repeated 4 times. Stage 2: predict the transmission layer. CBAM: Convolutional Block Attention Modules~\cite{woo2018cbam}. The output transmission image at iteration $i-1$ will be fed back to the network as the input of iteration $i$, and $\mathbf{\hat{T}}_0$ is initialized as $\mathbf{I}$, $1/N$ indicates the scaling ratio, \ie, H / N $\times$ W / N,  where H, W are the height and width of the input image.}
\label{fig:framework}
\vspace{-13pt}
\end{figure*}

\section{Related Work}

A variety of reflection removal methods, such as multi-view or video-based~\cite{gai2011blind, guo2014robust, li2013exploiting, nandoriya2017video, niklaus2021learned, sarel2004separating, sinha2012image, sun2016automatic, szeliski2000layer, xue2015computational}, dual-pixel sensor\cmx{-based}~\cite{punnappurath2019reflection} and polarization-based~\cite{chang2018single, lei2020polarized, simon2015reflection} methods, have been proposed to restore the transmission layer through motion or optical cues. Since we focus on SIRR in this paper, we mainly review the \cmx{single image based methods below.} 

To handle the ill-posed SIRR problem, traditional optimization-based methods introduce different priors~\cite{Levin2007,levin2003learning}. Observing that reflection layers are usually out of focus and appear to be more blurry than transmission layers, Li~\etal~\cite{li2014single} introduced a relative smoothness prior to distinguish the gradients of the two layers with different probability distributions. Shih~\etal~\cite{shih2015reflection} exploited ghosting cues to remove reflections when the thickness of the glass cannot be ignored. Huang~\etal~\cite{huang2019removing} proposed a wavelet transform based regularization \cmx{method} to separate ghosting patterns from background patterns. 
\cmx{Multi-scale depth-of-field (DoF) analysis based methods~\cite{wan2018region, wan2016depth} were proposed to separate reflection from transmissions by detecting the reflection-dominated regions.}
%
However, the thresholds in these methods for determining the reflection regions are vulnerable to noise.
In~\cite{arvanitopoulos2017single,yang2019fast}, the Laplacian data fidelity term is used to suppress the blurry reflections. However, these two methods cannot effectively remove strong reflections and might smooth out the transmission layer's details. In contrast, we leverage Laplacian features to emphasize the boundaries of strong reflections as a clue for the network to remove them.



Recently, \cmx{many} deep-learning-based methods~\cite{fan2017generic, jin2018learning, kim2020single, li2020single, wei2019single, wen2019single, yang2018seeing, zhang2019fast, zhang2018single} \cmx{were proposed} to solve the SIRR problem by learning task-specific features. 
Fan~\etal~\cite{fan2017generic} designed a deep neural network, \cmx{called} CEIL-Net, to first regress the edge map of the transmission layer and then reconstruct the transmission layer. Yang~\etal~\cite{yang2018seeing} \cmx{proposed the BDN}, a multi-stage network for estimating two layers sequentially, where the reflection layer predicted in the previous stage is used as auxiliary information to guide the transmission layer reconstruction in the next stage.
Li~\etal~\cite{li2020single} \cmx{proposed the IBCLN method}, which is an LSTM based recurrent network for concurrently refining the results of the predicted reflection and transmission layers.
Wan~\etal~\cite{wan2019corrn} \cmx{proposed} a feature-sharing strategy and a statistical loss for removing the strong reflections within local regions. \cmx{They further proposed to study the face reflection removal problem in~\cite{wan2020face}.}
%
%
\cmx{Zhang~\etal~\cite{zhang2019fast} proposed to explore the edge hints in the user-specified regions to separate the reflection and transmission layers.} %
\cmx{Some methods proposed to generate more representative training data. 
Jin~\etal~\cite{jin2018learning} proposed multiple data generation models. Wen~\etal~\cite{wen2019single} proposed SynNet to generate images with reflections beyond linearity. Wei~\etal~\cite{wei2019single} introduced an alignment-invariant loss to utilize the misaligned images as the real-world training dataset. Kim~\etal~\cite{kim2020single} proposed a physics-based rendering method to render realistic images with reflections.}
%
%
\cmx{Unlike these methods, this paper explores how to incorporate the location information of reflections into the network for controlling the feature flow.} 

\cmx{Many loss terms were used to boost the SIRR performance~\cite{kim2020single, li2020single, wei2019single, wen2019single, yang2018seeing, zhang2018single}, \eg, the VGG-based perceptual loss, exclusion loss in the gradient domain, and adversarial loss to prevent the blurring effects. 
In this paper, a new composition loss without using the GT reflection-dominated region masks is proposed to train our RDM.}
\vspace{-3pt}
\section{Method}
\label{sec:method}

\begin{figure*}[t]
\centering
\includegraphics[width=\linewidth]{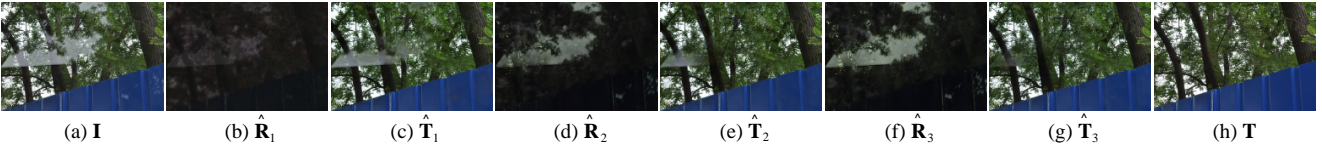}
\caption{The gradual refinement of reflection removal results after each iteration. The input image $\mathbf{I}$ is taken in front of a window glass.}
\label{fig:iteration}
\vspace{-12pt}
\end{figure*}
Our network is a recurrent network, as illustrated in Fig.~\ref{fig:framework}. In each iteration $i$, our network takes the original image $\mathbf{I}$ and the transmission layer $\mathbf{\hat{T}}_{i-1}$ predicted in the previous iteration $i-1$ as inputs, and predicts the transmission layer $\mathbf{\hat{T}}_i$ to continue the iteration. $\mathbf{\hat{T}}_0$ is initialized as $\mathbf{I}$. \dz{The recurrent structure of our network is inspired by IBCLN~\cite{li2020single}. However, different from jointly estimating the transmission and reflection layers in IBCLN, we design our model to reconstruct the transmission layer $\mathbf{\hat{T}}_{i}$ conditioned on the RCMap and the restored reflection layer at each iteration.} The step-by-step refinement results of reflection removal are shown in Fig.~\ref{fig:iteration}. 
%


%

\dz{Like BDN~\cite{yang2018seeing}, each iteration of our network is divided into two stages to restore two layers sequentially. 
However, we leverage RCMap to control the between-stage information flow.
In the first stage, we predict the reflection layer $\mathbf{\hat{R}}_{i}$ and RCMap $\mathbf{\hat{C}}_{i}$ by taking $\mathbf{I}$ and $\mathbf{\hat{T}}_{i - 1}$ as inputs. We denote the first stage as the function $G_R$:}
{
\setlength\abovedisplayskip{5pt}
\setlength\belowdisplayskip{5pt}
\begin{equation}
\mathbf{\hat{R}}_{i}, \mathbf{\hat{C}}_{i} = G_R(\mathbf{I}, \mathbf{\hat{T}}_{i-1}).
\end{equation}\label{eqn:stage1}%
This stage mainly consists of two modules: the reflection detection module (RDM) and the transmission-feature suppression module (TSM). Specifically, The RDM takes $\mathbf{I}$ and $\mathbf{\hat{T}}_{i-1}$ as inputs and predicts the confidence map $\mathbf{\hat{C}}_{i}$ using features from a multi-scale Laplacian sub-module (MLSM). Next, the TSM is used to suppress the Laplacian features within transmission-dominated regions via an element-wise multiplication between the features and $\mathbf{\hat{C}}_{i}$. Afterwards, the suppressed features and the image features are concatenated as the inputs of an LSTM~\cite{hochreiter1997long} block to estimate $\mathbf{\hat{R}}_i$. In our work, $\mathbf{\hat{R}}_i$ and $\mathbf{\hat{C}}_i$ are mainly used as cues to facilitate the reconstruction of transmission layers.}

In the second stage, we predict the transmission layer $\mathbf{\hat{T}}_{i}$ with the input of $\mathbf{I}, \mathbf{\hat{T}}_{i-1}$ as well as $\mathbf{\hat{R}}_i, 1 - \mathbf{\hat{C}}_i$ computed in the first stage. We denote the second stage as a function $G_T$ which can be described as:
{
\setlength\abovedisplayskip{5pt}
\setlength\belowdisplayskip{5pt}
\begin{equation}
\mathbf{\hat{T}}_{i} = G_T(\mathbf{I}, \mathbf{\hat{T}}_{i-1}, \mathbf{\hat{R}}_{i}, 1 - \mathbf{\hat{C}}_{i}).
\end{equation}
Notice that we utilize the inverse confidence map, \ie $1 - \mathbf{\hat{C}}_i$, in this stage. Since the transmission layer dominates at regions where $1 - \mathbf{\hat{C}}_i$ have a high value,
we expect this map to help the network learn weights to encode the reflection information in an adaptive manner, which should benefit the reconstruction of the transmission layer. For the network structure in this stage, we follow the contextual auto-encoder network in~\cite{qian2018attentive}, and additionally leverage CBAM (Convolutional Block Attention Module)~\cite{woo2018cbam} blocks after \textit{Conv} and \textit{ReLU} to compute the channel-wise and spatial attention. Please refer to Sec.1 in the supplemental material for detailed network architecture.}




\begin{figure}[t]
\centering
\includegraphics[width=\linewidth]{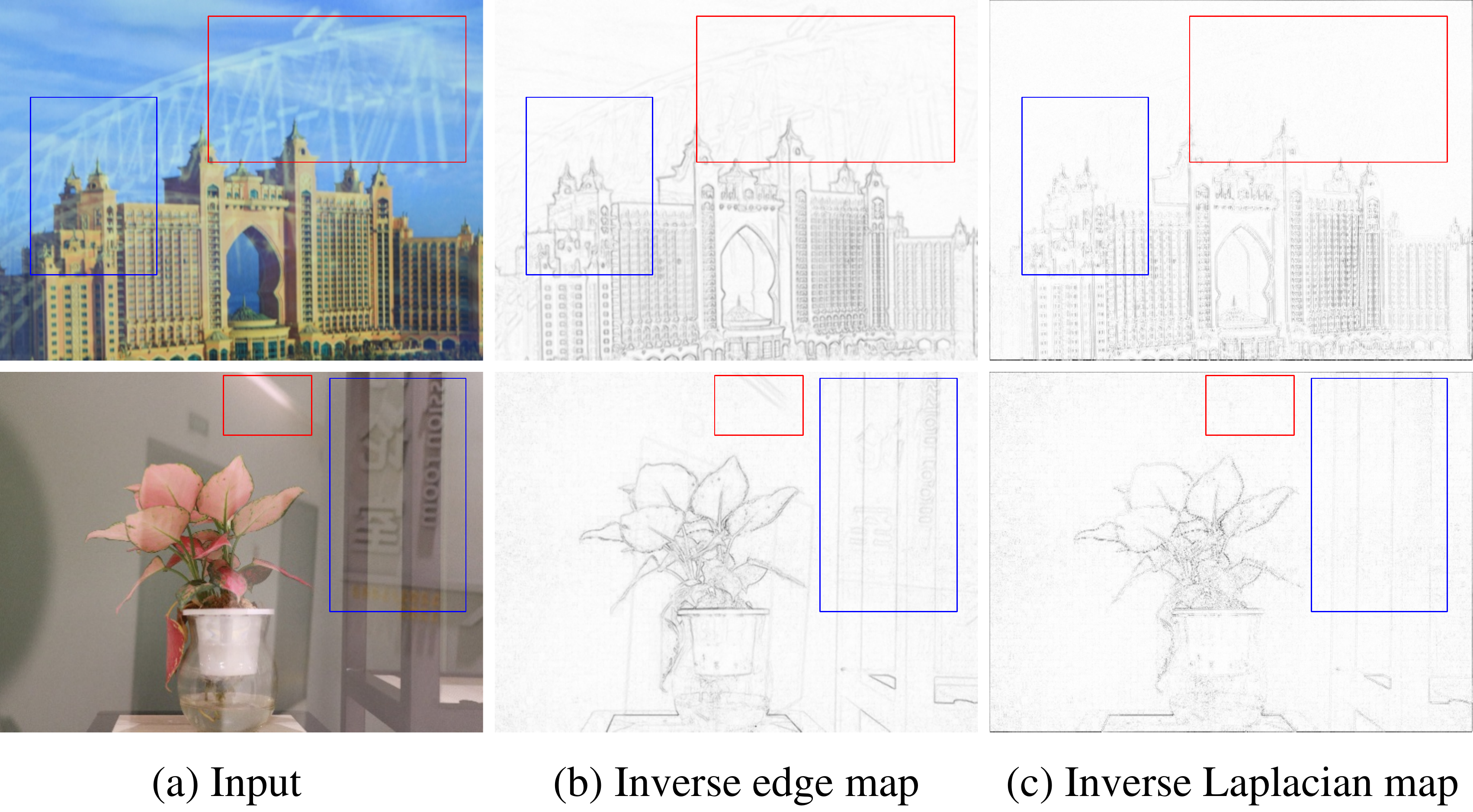}
\caption{\cmx{Two inverse edge / Laplacian maps along with the input images. We compute edge map $\mathbf{E}$ using the method in~\cite{fan2017generic} to obtain an image ranging between $0$ and $1$. Similarly, we compute absolute Laplacian values through convolution with kernel $\mathbf{k}_L$, then normalize each Laplacian value to obtain a map $\mathbf{L}$. However, we compute two inverse maps, \ie, $1 - \mathbf{E}$, $1 - \mathbf{L}$ for better visualization. Thus, value $0$ in the gradient domain is mapped to $1$ in the inverse maps.}}
\label{fig:laplacian_maps}
\vspace{-16pt}
\end{figure}

\noindent\textbf{Multi-scale Laplacian Features.}\quad We observe that the Laplacian operator, a second-order differential operator, can suppress the low-frequency reflections better. As illustrated in Fig.~\ref{fig:laplacian_maps}, low-frequency reflections are less evident in the inverse Laplacian map than in the inverse edge map, which suggests that the Laplacian operator suppresses low-frequency reflections more effectively. In contrast, strong reflections that have hard boundaries can not be suppressed by the Laplacian operator. It is possible that the difference between $\mathbf{I}$ and $\mathbf{T}$ caused by strong reflections becomes more obvious in the Laplacian domain. We assume such a behavior of the Laplacian operator is beneficial to detect reflection-dominated regions and thus concatenate these two images to form $\mathbf{X}_{in}=[\mathbf{I},\mathbf{\hat{T}}_{i-1}]$ as inputs to obtain multi-scale Laplacian features. 

\begin{figure}[t]
\centering
\includegraphics[width=\columnwidth]{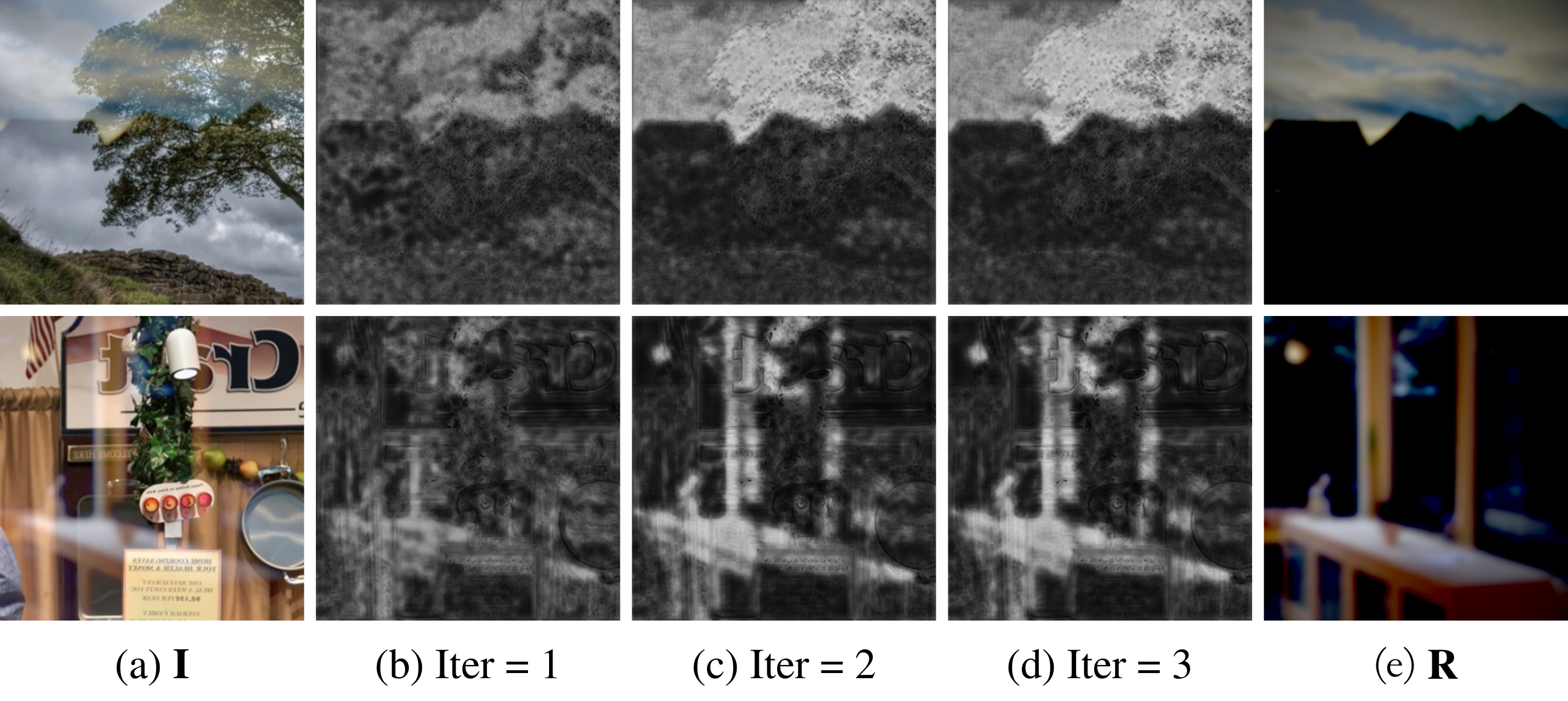}
\caption{Improvement of the RCMaps during iterations. Note that the reflection-dominated regions are gradually evident and accurate. Refer to Sec.~9 of the supplemental material for more results.}
\label{fig:confidence_map}
\vspace{-18pt}
\end{figure}


For the purpose of multi-scale Laplacian feature learning, \dz{we down-sample $\mathbf{X}_{in}$ through scaling the height and width of input images using bi-linear interpolation with the following factors: 1/2, 1/4, 1/8~\cite{gonzalez2004digital}}. The down-sampled results are denoted by $\mathbf{X}_{2}^{\downarrow}, \mathbf{X}_{4}^{\downarrow}, \mathbf{X}_{8}^{\downarrow}$ respectively. We utilize a convolution kernel with weights initialized to be a $3 \times 3$ Laplacian kernel, denoted by $\mathbf{k}_L = [0, -1, 0; -1, 4, -1; 0, -1, 0]$, to obtain the second derivative signal from $\mathbf{X}_{in}$. We allow the network to fine-tune the Laplacian kernel parameters to better extract Laplacian features, where the fine-tuned parameters are denoted as $\mathcal{L}ap$. During training, we utilize gradient clipping (0.25 in our experiments) to make sure that the learned kernel stays close to the original kernel $\mathbf{k}_L$.

After the Laplacian convolution block in Fig.~\ref{fig:framework}, the up-sampling operation $U^{\uparrow}_j$ is applied to the multi-scale Laplacian feature maps to restore their original size, where $j$ is the sampling rate. Precisely, given the images $\mathbf{X}_{j}^{\downarrow}(j=2,4,8)$ and $\mathbf{X}_{in}$, the output features $\mathbf{X}_{out}$ can be written as:
{
\setlength\abovedisplayskip{5pt}
\setlength\belowdisplayskip{5pt}
\begin{equation}
\mathbf{X}_{out}=Concat(\mathcal{L}ap(\mathbf{X}_{in}), U_j^{\uparrow}(\mathcal{L}ap(\mathbf{X}^{\downarrow}_{j})))_{j=2,4,8},
\end{equation}
\noindent{where} $Concat$ is the concatenation operation.}

Finally, taking $\mathbf{X}_{out}$ as input, RDM predicts the reflection confidence map $\mathbf{\hat{C}}$ from the Laplacian features. We employ three Squeeze-and-Excitation Residual Block (SE-ResBlocks)~\cite{hu2018squeeze} to get efficient multi-channel Laplacian features, where each block comprises of three layers of SE-ResNet, then the PReLU function~\cite{he2015delving} is used to activate the features while keeping the negative values. These combined blocks denotes as $f_{\mathcal{L}ap}$. Thus, given the features $\mathbf{X}_{out}$, the map $\mathbf{\hat{C}}$ can be described as:
\setlength\abovedisplayskip{5pt}
\setlength\belowdisplayskip{5pt}
\begin{equation}
\mathbf{\hat{C}} = Sigmoid(Conv(f_{\mathcal{L}ap}(\mathbf{X}_{out}))).
\end{equation}
The improvement of the predicted RCMaps for two synthesized images along with iterations is illustrated in Fig.~\ref{fig:confidence_map}.


\noindent\textbf{Transmission-feature suppression module.}\quad In this module, we employ three SE-ResBlocks to refine the Laplacian features and then multiply the features by $\mathbf{\hat{C}}_i$ for the purpose of transmission features suppressing. Instead of predicting $\mathbf{\hat{R}}_{i}$ with RCMap $\mathbf{\hat{C}}_i$ as input, just like predicting $\mathbf{\hat{T}}_{i}$ in the second stage, we empirically found that suppressing the part of Laplacian features belong to transmission-dominated regions benefits the reflection layer prediction. It also leads to a relatively simple network design by concatenating the features computed with $\mathbf{X}_{in}$ and the suppressed Laplacian features as the input to the LSTM~\cite{hochreiter1997long} block. That is, the same encoder-decoder network structure in the second stage is not used in the first stage to reduce the number of network parameters.

\vspace{-5pt}
\section{Training Loss}
\label{sec:loss_function}

In this section, we describe the four loss functions used in the training of our network. For clarity, we denote the ground-truth transmission and reflection layers by $\mathbf{T}, \mathbf{R}$, the predicted transmission and reflection layers at iteration $i$ as $\mathbf{\hat{T}_i}, \mathbf{\hat{R}}_i$ respectively, \dz{and inverse gamma correction as a function $g_{inv}$}. The iterations number used in our recurrent network is denoted by $N$.

\noindent \textbf{Composition Loss.}\quad The composition loss is proposed to guide the training of RDM for predicting $\mathbf{\hat{C}}_i$ and supervise $\mathbf{\hat{T}}_i$, $\mathbf{\hat{R}}_i$ at each iteration using training images synthesized by the following linear alpha blending model in~\cite{zhang2018single}: {
\setlength\abovedisplayskip{5pt}
\setlength\belowdisplayskip{5pt}
\begin{equation}
\mathbf{\Tilde{I}}=\alpha \cdot \mathbf{\Tilde{T}}+\mathbf{\Tilde{R}}
\label{eq:linear_alpha_blending}
\end{equation}
where $\alpha$ is a scalar, $\mathbf{\Tilde{T}} = g_{inv}(\mathbf{T})$ and $\mathbf{\Tilde{R}} = g_{inv}(\mathbf{R})$.}

Firstly, since the map $1 - \mathbf{\hat{C}}_i$ can serve as the weight map $\mathbf{W}$ in Eq.~\ref{eq:reflection_decomposition}, we can compose an image $\mathbf{\hat{I}}_i$ by the following formula using gamma corrected $\mathbf{T}$ and $\mathbf{R}$: {
\setlength\abovedisplayskip{5pt}
\setlength\belowdisplayskip{5pt}
\begin{equation}
\label{eq:reflection_composition_2}
\mathbf{\hat{I}}_i=(1 - \mathbf{\hat{C}}_i)\circ \mathbf{T} + \mathbf{R},
\end{equation}
where $\circ$ is an element-wise production operation.}
We formulate the loss for RCMap as follows:
{
\setlength\abovedisplayskip{5pt}
\setlength\belowdisplayskip{5pt}
\begin{equation}
\label{eq:loss_c}
\mathcal{L}_{\mathbf{\hat{C}}} = \sum_{\mathbf{I, T, R} \in \mathcal{D}} {\sum_{i=1}^{N} \theta^{N-i}\mathcal{L}_{MSE}(\mathbf{I},\mathbf{\hat{I}}_i}),
\end{equation}
where $\mathcal{L}_{MSE}$ indicates the mean squared error, $\theta$ is an attenuation coefficient to indicate the strength of supervision and we set it to 0.85.}

\cm{Secondly, same as IBCLN~\cite{li2020single}, we adopt Eq.~\ref{eq:linear_alpha_blending} to form a residual loss to guide the prediction of $\mathbf{\hat{T}}_i, \mathbf{\hat{R}}_i$ in two forms: $\mathbf{\hat{I}}_i^{g}=\alpha \cdot g_{inv}(\mathbf{T}) + g_{inv}(\mathbf{\hat{R}}_i)$ and $\mathbf{\hat{I}}_i^g =\alpha \cdot g_{inv}(\mathbf{\hat{T}}_i) + g_{inv}(\mathbf{\hat{R}}_i)$, where $\alpha$ is a known scalar used to synthesize training images. We use Eq.~\ref{eq:loss_c} to compute the difference between $\mathbf{\hat{I}}_i^{g}$ and $g_{inv}(\mathbf{I})$, and denote the loss as $\mathcal{L}_{res}$. The composition loss for the synthesized images is defined as:}
\begin{equation}
\mathcal{L}_{comp} = \mathcal{L}_{\mathbf{\hat{C}}} + \mathcal{L}_{res}.
\end{equation}


\noindent\textbf{Perceptual Loss.}\quad 
We use VGG-19 network~\cite{simonyan2014very} pre-trained on ImageNet~\cite{russakovsky2015imagenet} dataset to extract features for the computation of our perceptual loss. This loss takes multi-scale images as inputs and can be written into:
{
\setlength\abovedisplayskip{5pt}
\setlength\belowdisplayskip{5pt}
\begin{equation}
\mathcal{L}_{p} = \sum_{\mathbf{T}^j\in \mathcal{D}}\sum_{j=1,3,5}\gamma_j\mathcal{L}_{VGG}(\mathbf{T}^j,\mathbf{\hat{T}}^j_N),
\end{equation}
where $\mathcal{L}_{VGG}$ denotes the $l_1$ loss
between VGG features. $\mathbf{\hat{T}}^j_N$ indicates the output of the last $j^{th}$ layer of the autoencoder in stage 2 at iteration N, and $\mathbf{T}^j$ indicates the ground truth that has the same scale as $\mathbf{\hat{T}}^j_N$. We set $\gamma_1=1, \gamma_3=0.8, \gamma_5=0.6$ respectively. For $\mathcal{L}_{VGG}$, we use the layers `conv$k$\_2' ($k=1, 2, 3, 4, 5$) of the standard VGG-19 net.
Fig.~\ref{fig:framework} shows how the network computes $\mathbf{\hat{T}}^j_N$.}

\noindent\textbf{Pixel and SSIM Loss.}\quad The pixel loss is used to penalize the pixel-wise difference between $\mathbf{T}$ and $\mathbf{\hat{T}}_i$. Here, we utilize $l_1$ norm loss, denoted as $\mathcal{L}_1$, to compute the absolute difference. We define the pixel loss as:
{
\setlength\abovedisplayskip{5pt}
\setlength\belowdisplayskip{5pt}
\begin{equation}
\mathcal{L}_{pixel} = \sum_{\mathbf{T} \in \mathcal{D}}\sum_{i=1}^N \theta^{N - i} \mathcal{L}_1(\mathbf{T}, \mathbf{\hat{T}}_i),
\label{eq:pixel_loss}
\end{equation}
where $\theta$ is set to 0.85 as well.}


It is verified that the SSIM(structural similarity index) loss combined with $l_1$ loss perform better than $l_2$ loss in image restoration~\cite{zhao2015loss}. Therefore, we also adopt $\mathcal{L}_i^{SSIM}= 1 - SSIM(\mathbf{T}, \mathbf{\hat{T}}_i)$ in each iteration $i$ as a loss term, which can be written into:
{
\setlength\abovedisplayskip{5pt}
\setlength\belowdisplayskip{5pt}
\begin{equation}
\mathcal{L}_{SSIM} = \sum_{\mathbf{T} \in \mathcal{D}}\sum_{i=1}^N \theta^{N - i}\mathcal{L}^{SSIM}_{i},
\label{eq:ssim_loss}
\end{equation}
where the setting of $\theta$ is same as Eq.~\ref{eq:pixel_loss}. We denote the mixture of SSIM and pixel loss as $\mathcal{L}_{mix}$ and define it as:}
\begin{equation}
\mathcal{L}_{mix} = \beta \mathcal{L}_{SSIM} + (1 - \beta) \mathcal{L}_{pixel},
\end{equation}
\noindent where $\beta$ is set to 0.84, following the design in~\cite{zhao2015loss}.

\noindent\textbf{Adversarial Loss.}\quad To improve the quality of the restored images, we further add an adversarial loss. We adopt a multi-layer discriminator network \textbf{\textit{D}} to assess the quality of images and define the adversarial loss as:
{
\setlength\abovedisplayskip{5pt}
\setlength\belowdisplayskip{5pt}
\begin{equation}
\mathcal{L}_{adv} = \sum_{\mathbf{T}\in\mathcal{D}} -log\ \textbf{\textit{D}}(\mathbf{T}, \mathbf{\hat{T}}).
\end{equation}
\textbf{Overall Loss.}\quad Totally, our training loss is defined as:} 
\begin{equation}
\mathcal{L}= \lambda_1 \mathcal{L}_{comp}+ \lambda_2 \mathcal{L}_{p} + \lambda_3 \mathcal{L}_{mix} + \lambda_4 \mathcal{L}_{adv}.
\end{equation}
We empirically set the weights for each loss in our experiments as: $\lambda_1 = 0.4, \lambda_2 = 0.2, \lambda_3 =0.4, \lambda_4=0.01$.

\renewcommand\arraystretch{1.05}
\begin{table*}[t]
\begin{center}
\resizebox{0.92\textwidth}{!}{
\begin{tabular}{*{10}{c}}
\toprule[1.1pt]
\multirow{2}*{Dataset\ (size)} & \multirow{2}*{Index\ ($\uparrow$)} & \multicolumn{8}{c}{Methods}\\


\cmidrule[0.5pt]{3-10} &
& Zhang \etal.-F~\cite{zhang2018single} & BDN~\cite{yang2018seeing} & RMNet~\cite{wen2019single} & ERRNet-F~\cite{wei2019single} & CoRRN-F~\cite{wan2019corrn} & Kim \etal.~\cite{kim2020single} & IBCLN-F~\cite{li2020single} & Ours \\

\hline

\multirow{2}*{Postcard\ (199)} & PSNR  & 21.497 & 20.460 & 19.833 & 22.374 & 20.866 & 23.055 & \color{blue}{23.421} & \textbf{\color{red}{23.724}}\\
& SSIM  & 0.870 & 0.858 & 0.872 & \color{blue}{0.889} & 0.866 & 0.871 & 0.864 & \textbf{\color{red}{0.903}} \\
\hline

\multirow{2}*{Object\ (200)} & PSNR & 23.675 & 22.642 & 24.045 & 23.101 & \textbf{\color{red}{25.134}} & 23.552 & \color{blue}{24.416} & 24.361\\
& SSIM &  0.885 & 0.857 & 0.847 & 0.876 & \textbf{\color{red}{0.912}} & 0.879 & 0.889 & \color{blue}{0.898} \\

\hline

\multirow{2}*{Wild(55)} & PSNR & 24.861 & 22.048 & 19.800 & 24.097 & 24.341 & \color{blue}{25.534} & 24.724 & \textbf{\color{red}{25.731}}\\
& SSIM & 0.886 & 0.828 & 0.885 & 0.880 & \color{blue}{0.893} & 0.890 & 0.871 & \textbf{\color{red}{0.902}} \\

\hline

\multirow{2}*{Zhang \etal.(20)} & PSNR & 22.230 & 18.487 & 18.780 & \color{blue}{23.153} & 21.569 & 20.218 & 21.008 & \textbf{\color{red}{23.338}}\\ & SSIM & 0.800 & 0.729 & 0.708 & \color{blue}{0.809} & 0.807 & 0.735 & 0.760 & \textbf{\color{red}{0.812}} \\

\hline

\multirow{2}*{Li \etal.(20)} & PSNR &  20.721 & 18.828 & 15.457 & 20.368 & 21.841 & 20.096 & \textbf{\color{red}{23.695}} & \color{blue}{23.451}\\ & SSIM &  0.765 & 0.738 & 0.732 & 0.771 & \color{blue}{0.805} & 0.759 & 0.804 & \textbf{\color{red}{0.808}} \\

\hline

\multirow{2}*{\textit{Average}(494)} & PSNR & 22.752 & 21.374 & 21.315 & 22.810 & 23.049 & 23.298 & \color{blue}{23.882} & \textbf{\color{red}{24.179}}\\ & SSIM & 0.871 & 0.844 & 0.851 & 0.875 & \color{blue}{0.883} & 0.866 & 0.868 & \textbf{\color{red}{0.893}} \\

\bottomrule[1.5pt]
\end{tabular}}
\end{center}
\vspace{-7pt}
\caption{Quantitative comparisons to state-of-the-art methods on real-world datasets. The best results are marked in \textcolor{red}{red}, and the second-best results are marked in \textcolor{blue}{blue}. }
\label{tab:comparison}
\vspace{-13pt}
\end{table*}

\renewcommand{\thefootnote}{1}
\section{Experiments}
\label{sec:experiments}
We implement our method using PyTorch~\cite{paszke2017automatic} on a PC with an Nvidia Geforce RTX 2080 Ti GPU. \cmx{To minimize the training loss, we adopt ADAM optimizer~\cite{kingma2014adam} to train our network for 60 epochs with a learning rate $2e^{-4}$ and batch size $2$. After 60 epochs, we reduce the learning rate to $1e^{-4}$ and add the unaligned dataset from ERRNet~\cite{wei2019single} to fine-tune our model.} The $\beta_1$, $\beta_2$ in ADAM are set to 0.5 and 0.99, respectively. The network weights are initialized using a normal distribution (mean:0, variance: 0.02), and the iteration number $N$ is set to 3, same as IBCLN~\cite{li2020single}.  Our network has 10.926M parameters, its FLOPs are 111.63G, which is comparable to IBCLN (130.88G) at each iteration. In the inference stage, it takes about $0.068s$ to process an input image of resolution $400 \times 540$.
%

\noindent\textbf{Training dataset.}\quad Our training dataset consists of both synthetic and real-world data. For the synthetic data, we use the images dataset from~\cite{fan2017generic}. \dz{This dataset has approximately 13700 image pairs of size $256\times 256$. With these pairs, we adopt Eq.~\ref{eq:linear_alpha_blending} and randomly sample $\alpha$ in $[0.8,1.0]$ to obtain $\mathbf{\Tilde{I}}$. We then apply gamma correction~\cite{yaroslavsky2012digital} to $\{\mathbf{\Tilde{I}}, \mathbf{\Tilde{T}}, \mathbf{\Tilde{R}}\}$ to generate image triples $\{\mathbf{I}, \mathbf{T}, \mathbf{R}\}$. For the real-world data, there are a total of 540 image pairs, $\{\mathbf{I}, \mathbf{T}\}$, in our dataset, including 200 pairs provided by the ``Nature'' dataset in IBCLN~\cite{li2020single}, 90 pairs provided by Zhang \etal~\cite{zhang2018single} and 250 pairs provided by the unaligned dataset in ERRNet~\cite{wei2019single}. \dz{Following IBCLN ~\cite{li2020single}}, we feed the network with 4000 pairs (triples) in each epoch, including 2800 triples randomly sampled from the synthetic data and 1200 pairs of size $256\times 256$ cropped from the real-world data. Moreover, our data augmentation can generate training images to cover more real-world reflection types. The details can be found in Sec.2 of the supplemental material.}



\subsection{Comparisons}
To evaluate the performance of our method, we compare it to seven state-of-the-art SIRR methods, including Zhang \etal.~\cite{zhang2018single}, BDN~\cite{yang2018seeing}, RMNet~\cite{wen2019single}, ERRNet~\cite{wei2019single}, CoRRN~\cite{wan2019corrn}, Kim \etal.~\cite{kim2020single}, and IBCLN~\cite{li2020single}. We use PSNR and SSIM as metrics, where the higher metric value means better performance. For fair comparisons, we report their better performances either using their original trained models or using models fine-tuned with our training dataset if their training codes are available. The fine-tuned results are denoted with a suffix "-F".
%
%
Note that we do not fine-tune the RMNet~\cite{wen2019single}, as it requires additional alpha blending masks from a SynNet~\cite{wen2019single}.
%
We also modify the code of Kim~\etal.~\cite{kim2020single} to compute SSIM in RGB space.

\noindent{\textbf{Quantitative comparisons.}}\quad Tab.~\ref{tab:comparison} reports the performance comparisons on five real-world datasets.
Datasets in the first three rows are all from $SIR^{2}$ constructed in~\cite{2017Benchmarking}, and the rest two datasets are from the evaluation set in Zhang~\etal~\cite{zhang2018single} and the ``Nature'' test dataset in Li~\etal~\cite{li2020single} respectively. 
%
%
It can be seen that our method is ranked as top-1 on the Postcard, Wild, and Zhang~\etal datasets, top-2 on the Object (SSIM ranking) and Li~\etal (PSNR ranking) datasets. Moreover, our method achieves the best average PSNR and SSIM scores. This verifies that our method can achieve superior performance in various real-world scenarios.

\begin{figure*}[t]
\centering
\includegraphics[width=\linewidth]{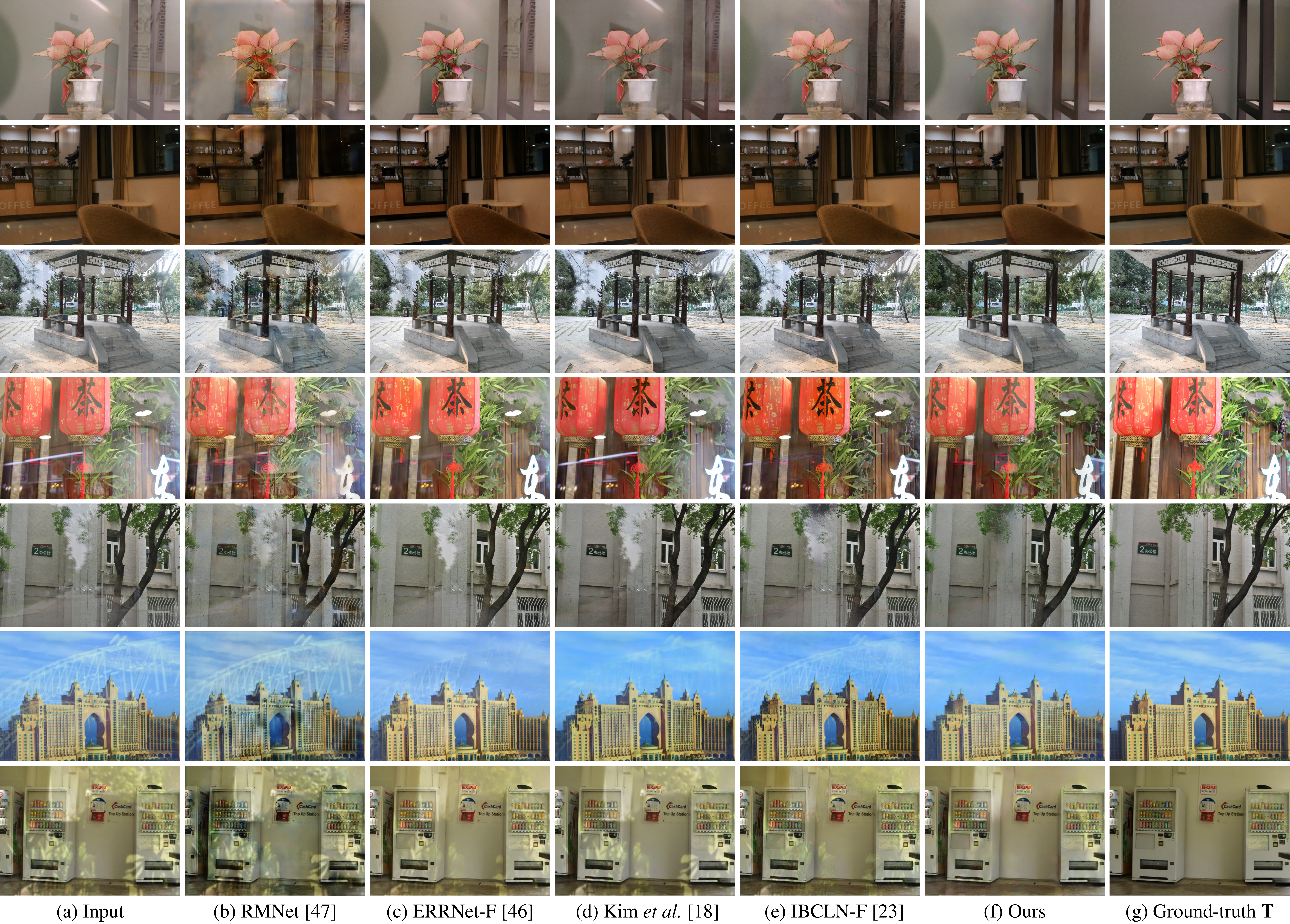}
\caption{Qualitative comparisons between the proposed method and four latest state-of-the-arts.}
\label{fig:visual_results}
\vspace{-15pt}
\end{figure*}
\noindent\textbf{Qualitative comparisons.}\quad  Fig.~\ref{fig:visual_results} shows the reflection removal results of four existing models and ours. These images are from the ``Nature'' test dataset of IBCLN~\cite{li2020single}(rows 1-2), unaligned datasets of ERRNet~\cite{wei2019single}(rows 3-5), and the benchmark datasets $SIR^{2}$~\cite{2017Benchmarking}(rows 6-7). It can be seen that existing methods typically fail to remove large-area reflections and strong highlights. In contrast, our method can remove most undesirable reflections while preserving high-frequency details in the transmission layer. More qualitative results can be found in our supplemental material.

\begin{figure}[t]
\centering
\includegraphics[width=\columnwidth]{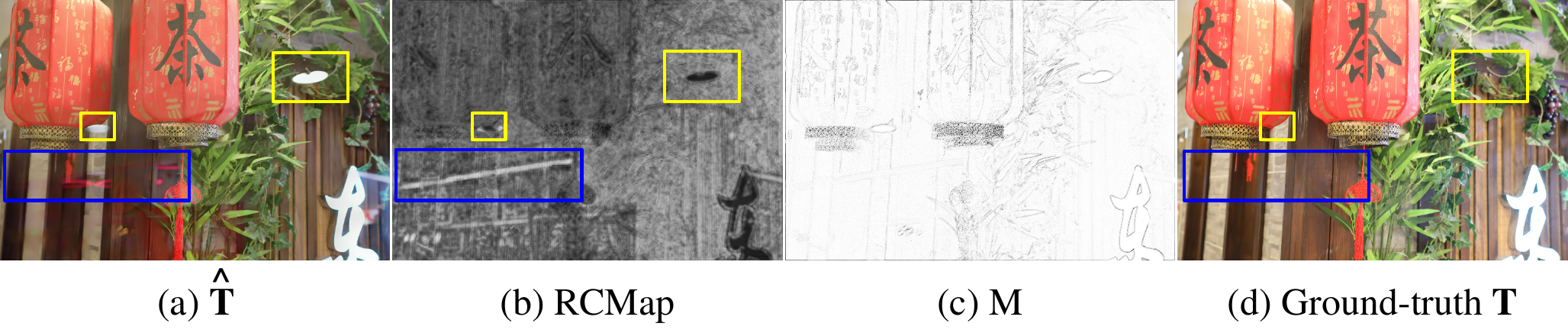}
\caption{A failure case. M: the inverse Laplacian map using learned Laplacian kernel. The reflected oval highlights are not correctly detected for the corrupted image in the fourth row of Fig.~\ref{fig:visual_results}.}
\label{fig:analysis_images}
\vspace{-17pt}
\end{figure}

However, when the strong reflection regions are not correctly detected, our method still lacks cues to remove such reflections. For instance, there are remained highlights in the 4th row of our results in Fig.~\ref{fig:visual_results}. As shown in Fig.~\ref{fig:analysis_images}, compared to the detected rod-shaped reflection (blue boxes), the two oval highlights (yellow boxes) are not completely detected. Hence, they still appear in the removal result. Although hard boundaries for these highlights exist in the inverse Laplacian map, our network still lacks enough context information to classify such oval highlights as reflections.

\subsection{Ablation Study}
\label{sec.ablation_study}
\vspace{-2pt}

To better analyze our network's architecture and evaluate the importance of the loss functions, we perform ablation studies by manipulating the model components, removing or replacing loss functions. The statistics of PSNR and SSIM are obtained by evaluating the re-trained models in these experiments.

\begin{figure*}[t]
\centering
\includegraphics[width=\linewidth]{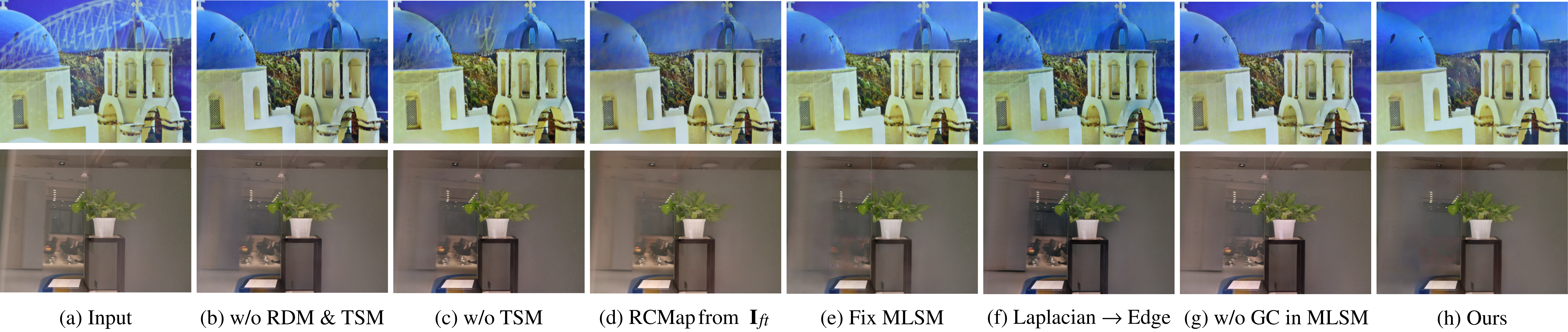}
\caption{The visualization of reflection removal results according to the ablation study in Tab.~\ref{tab:model_ablation}.}
\label{fig:ablation_figure}
\vspace{-15pt}
\end{figure*}

\noindent\textbf{Evaluation of the network architecture.}\quad \dz{In Tab.~\ref{tab:model_ablation}, we first show that three modules in our network, namely RDM, TSM, and LSTM, all contribute to the SIRR performance (first three rows).} We then test six different choices in the design of RDM, and the results are shown in the last six rows in Tab.~\ref{tab:model_ablation}. Our current design choices of RDM lead to the highest PSNR and SSIM scores. In addition, Fig.~\ref{fig:ablation_figure} illustrates the visual results of the ablation studies in Tab.~\ref{tab:model_ablation}.

\renewcommand\arraystretch{1}
\begin{table}[t]
\begin{center}
\resizebox{0.95\columnwidth}{!}{
\begin{tabular}{ccc|cc|cc}
\toprule[1.1pt]
\multirow{2}*{Model} & \multicolumn{2}{c}{\textit{SIR}$^2$~\cite{2017Benchmarking}} & \multicolumn{2}{c}{Zhang \etal.\cite{zhang2018single}} & \multicolumn{2}{c}{Li \etal.\cite{li2020single}}\\


\cmidrule[0.5pt]{2-7}
& PSNR & SSIM & PSNR & SSIM & PSNR & SSIM\\
\hline  

w/o RDM \& TSM  & 23.237 & 0.881 & 22.784 & 0.800 & 22.607 & 0.778\\

w/o TSM & 23.304 & 0.882 & 22.139 & 0.805 & 23.088 & 0.799\\

w/o LSTM & 23.808 & 0.887 & 21.040 & 0.760 & 22.721 & 0.794 \\

$\mathbf{\hat{C}}$ from $\mathbf{I}_{ft}$ & 22.595 & 0.888 & 21.747 & 0.792 & 22.283 & 0.792\\


MLSM $\rightarrow$ SLSM & 23.089 & 0.879 & 21.708 & 0.796 & 23.065 & 0.800\\

Fix MLSM & 23.640 & 0.891 & 22.951 & 0.809 & 22.646 & 0.802\\

$Laplacian \rightarrow Edge$ & 23.612 & 0.892 & 22.381 & 0.808 & 23.234 & 0.798\\

LKI $\rightarrow$ RKI & 24.106 & 0.893 & 21.784 & 0.791 & 22.224 & 0.794 \\

w/o GC in MLSM & 23.901 & 0.891 & 23.051 & \textbf{0.812} & 23.300 & 0.806 \\

\hline

Ours & \textbf{24.117} & \textbf{0.901} & \textbf{23.338} & \textbf{0.812} & \textbf{23.451} & \textbf{0.808}\\

\bottomrule[1.5pt]
\end{tabular}}
\end{center}
\vspace{-5pt}
\caption{\dz{Network structure ablation study. w/o $*$: remove module $*$. $M_1$ $\rightarrow$ $M_2$: change $M_1$ to $M_2$. SLSM: Single-scale Laplacian submodule. GC: gradient clipping. $Edge$: edge features, namely the partial derivatives of the image with respect to $x,y$ (refer to Fig.~\ref{fig:laplacian_maps}).
$\mathbf{\hat{C}}$ from $\mathbf{I}_{ft}$: disable MLSM in RDM and predict RCMap using the extracted features of $[\mathbf{I}$, $\mathbf{\hat{T}}_i]$ before LSTM block. Fix MLSM: disable the fine-tuning of Laplacian kernel parameters.}}
\label{tab:model_ablation}
\vspace{-22pt}
\end{table}

\begin{figure}[ht]
\setlength{\tabcolsep}{0.7pt}
 \centering
    \begin{tabular}{cccc}
    \vspace{-3pt}
    \includegraphics[width=0.244\linewidth,keepaspectratio]{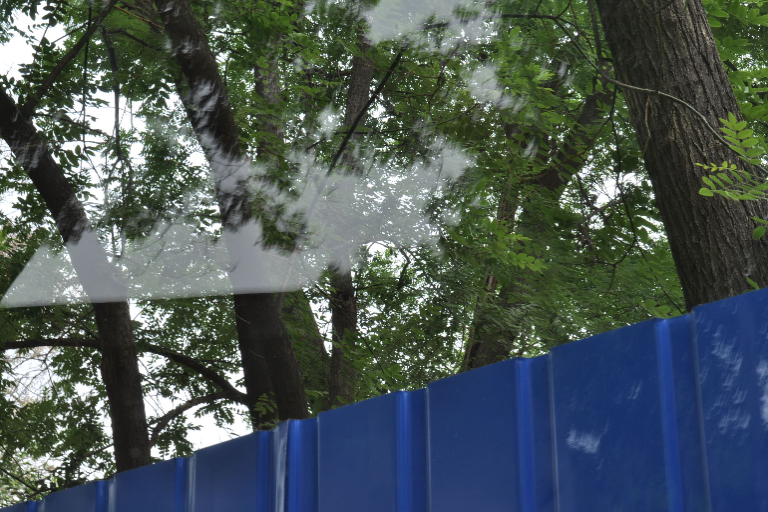}&  
    \includegraphics[width=0.244\linewidth,keepaspectratio]{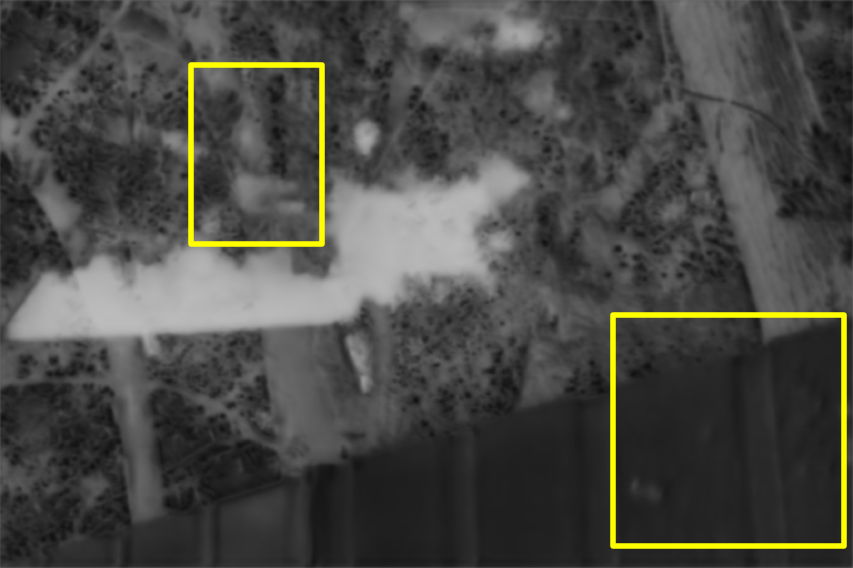}& 
    \includegraphics[width=0.244\linewidth,keepaspectratio]{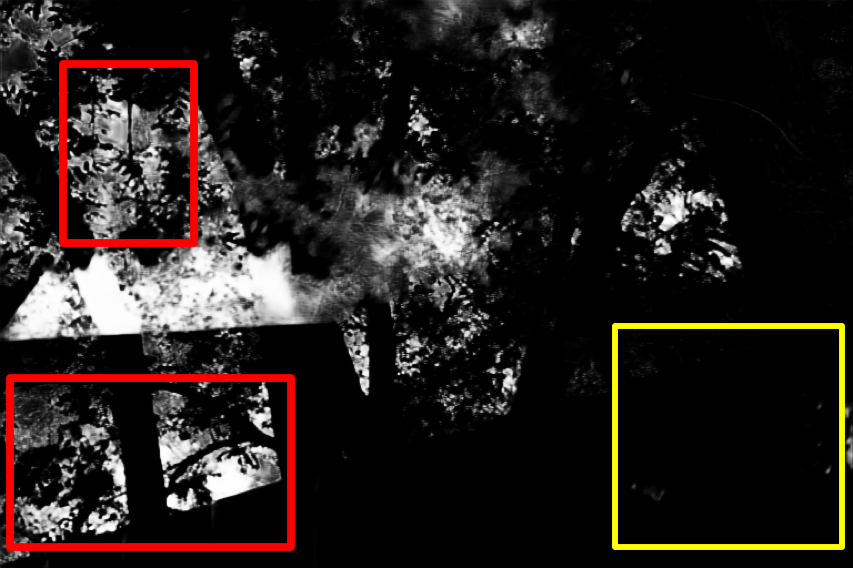}&  
    \includegraphics[width=0.244\linewidth,keepaspectratio]{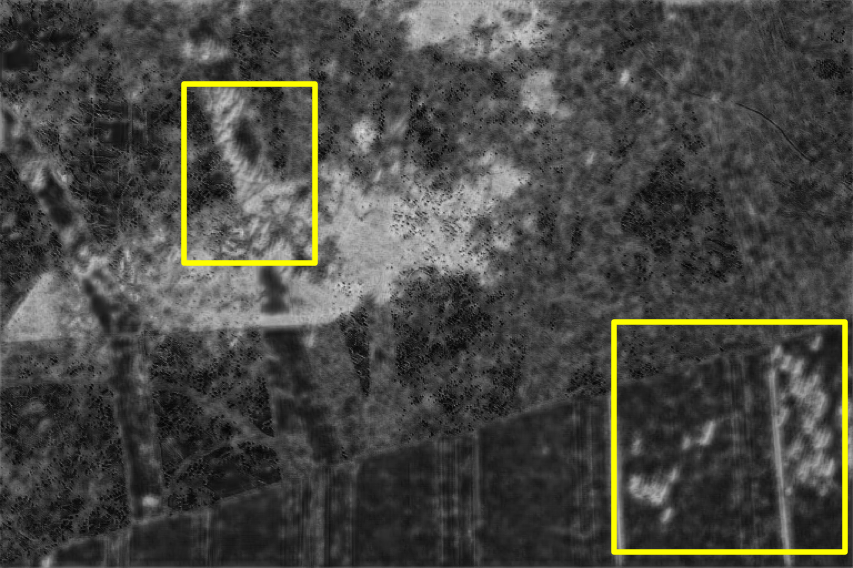} \\  
    \scriptsize (a) Input & 
    \scriptsize (b) RCMap [RKI] & 
    \scriptsize (c) RCMap [$\mathcal{L}_{\mathbf{\hat{C}}}^B$] &
    \scriptsize (d) RCMap [Ours] \\
    
    \vspace{-3pt}
    \includegraphics[width=0.244\linewidth,keepaspectratio]{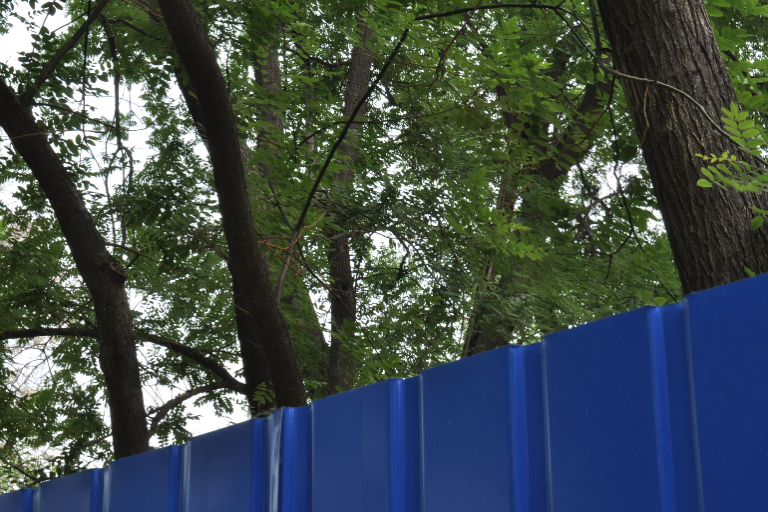}&
    \includegraphics[width=0.244\linewidth,keepaspectratio]{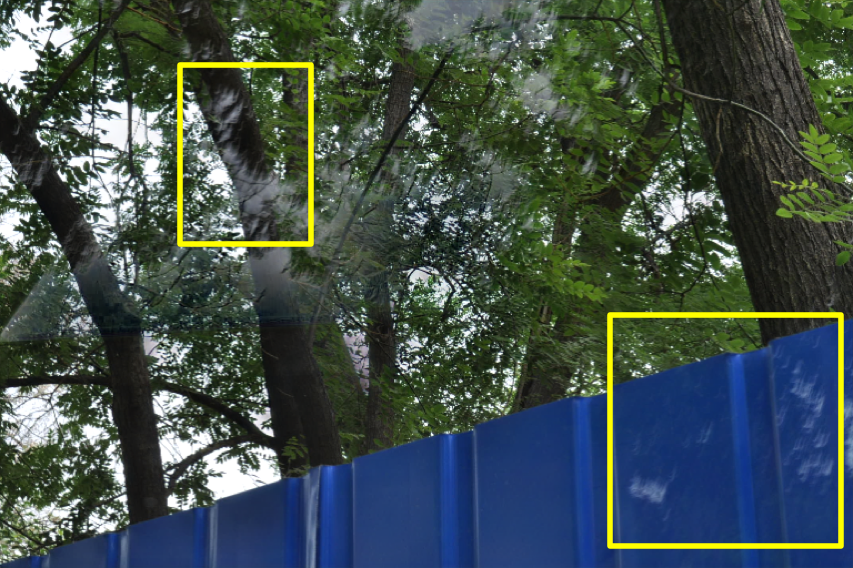}&
    \includegraphics[width=0.244\linewidth,keepaspectratio]{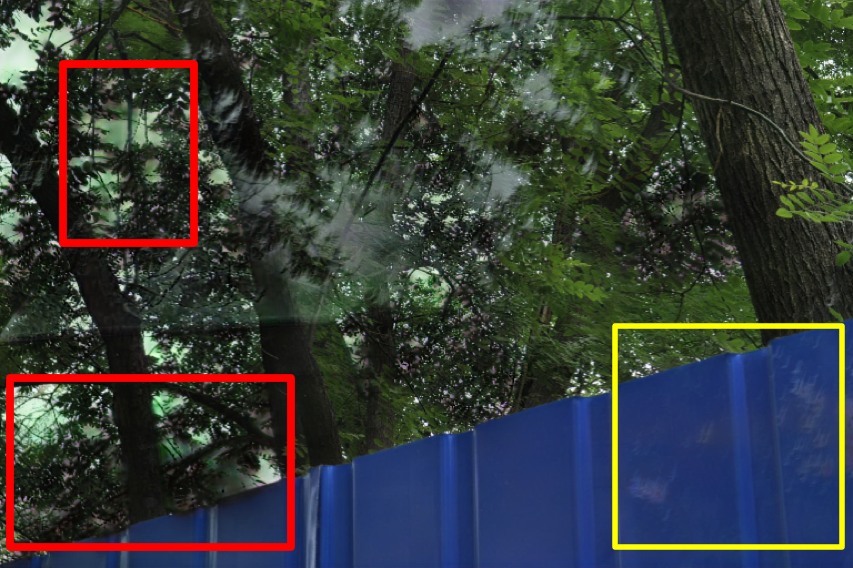}&
    \includegraphics[width=0.244\linewidth,keepaspectratio]{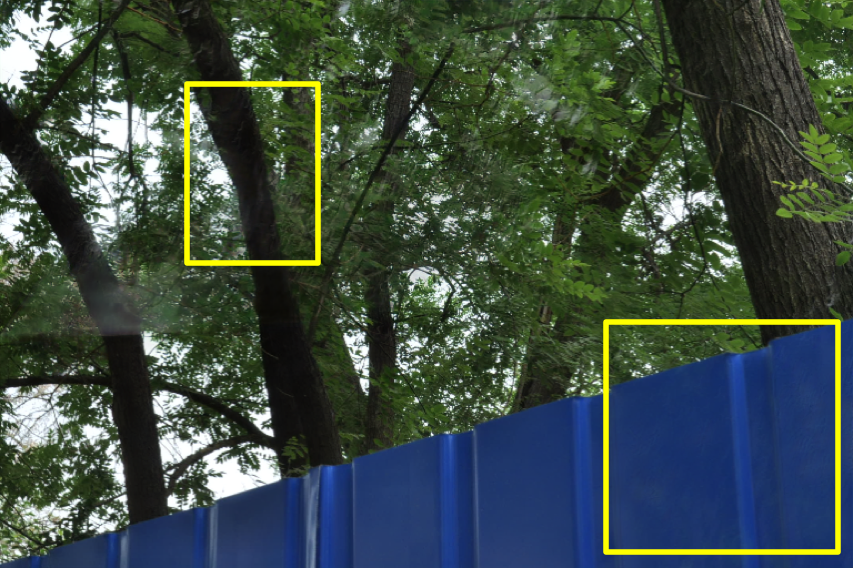} \\
    \scriptsize (e) Ground-truth $\mathbf{T}$ &
    \scriptsize (f) $\mathbf{T}$ [RKI] &
    \scriptsize (g) $\mathbf{T}$ [$\mathcal{L}_{\mathbf{\hat{C}}}^B$] &
    \scriptsize (h) $\mathbf{T}$ [Ours] \\
    
    \end{tabular}
    \caption{RCMaps and the predicted transmission images using RKI, $\mathcal{L}_{\mathbf{\hat{C}}}^B$ and our network.}
    \vspace{-11pt}
\label{fig:compar_rcmap}
\end{figure}

\begin{figure}[ht]
\setlength{\tabcolsep}{0.7pt}
 \centering
    \begin{tabular}{cccc}
    \vspace{-3pt}
    \includegraphics[width=0.244\linewidth,keepaspectratio]{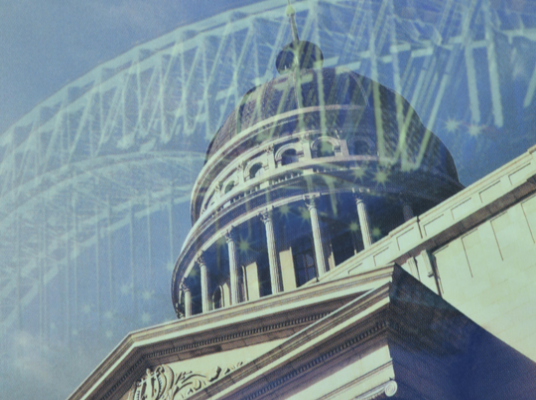}&  
    \includegraphics[width=0.244\linewidth,keepaspectratio]{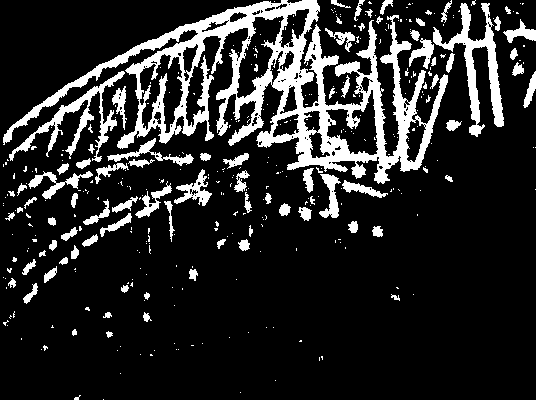}& %
    \includegraphics[width=0.244\linewidth,keepaspectratio]{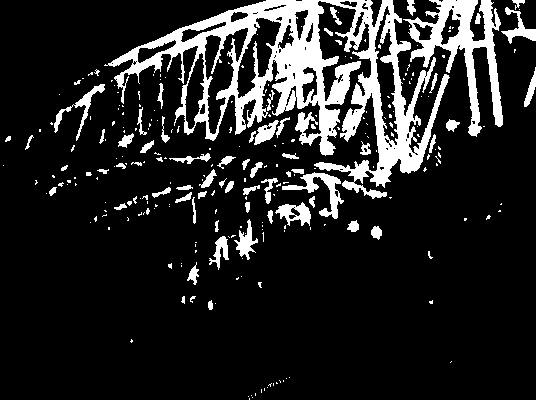} & \includegraphics[width=0.244\linewidth,keepaspectratio]{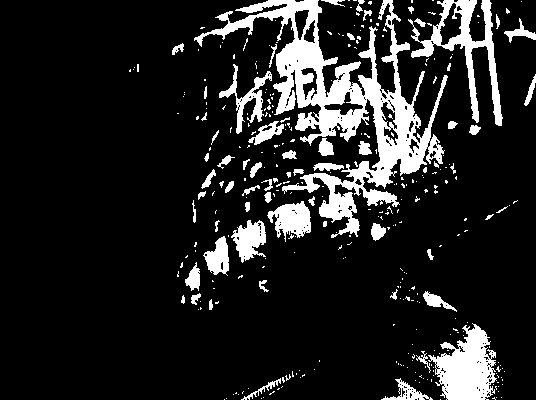}  \\
    \scriptsize (a) Input & 
    \scriptsize (b) Ours [th.=0.5] &
    \scriptsize (c) $\|\mathbf{I} - \mathbf{T}\|$ &
    \scriptsize (d) $\|\frac{\mathbf{I} - \mathbf{T}}{\mathbf{T}}\|$ \\
    
    \end{tabular}
    \caption{Calculated RCMaps. Ours~[th.=0.5]: apply a threshold (0.5) to our predicted RCMap.}
    \vspace{-18pt}
\label{fig:calculate_rcmap}
\end{figure}

Especially, to evaluate the effectiveness of Laplacian kernel initialization (LKI), we conduct an experiment by replacing LKI with random kernel initialization (RKI) using a Gaussian distribution (mean = 0, variance = 0.02) and cancel the gradient clipping. The PSNR/SSIM scores in the last second row of Tab.~\ref{tab:model_ablation} show that LKI is superior to RKI. Moreover, a qualitative comparison is shown in the second and fourth columns in Fig.~\ref{fig:compar_rcmap}. The kernel learned by RKI tends to smooth out small-size reflections (yellow boxes) in its RCMap, leading to a downgraded reflection-removal result. In contrast, ours can successfully detect and remove the reflections. Moreover, the kernels learned by LKI and RKI are available in Sec.3 of the supplemental material.

\noindent\textbf{Evaluation of loss functions.}\quad In Tab.~\ref{tab:loss_ablation}, we report the performance of our model re-trained with ablated loss terms (first five rows). It can be seen that each loss term contributes to the network's performance. We hypothesize that the performance drop after removing $\mathcal{L}_{SSIM}$ is due to the weight of SSIM loss is much larger than pixel loss in $\mathcal{L}_{mix}$.

\dz{
To verify the effectiveness of $\mathcal{L}_{\mathbf{\hat{C}}}$ in detecting reflection-dominated regions, we compare it with the raindrop removal method in ~\cite{qian2018attentive} that generates GT binary raindrop masks for the training.
Specifically, we replace the $\mathcal{L}_{\mathbf{\hat{C}}}$ with the combination of the binary cross-entropy (BCE) losses at different iterations:}
{
\setlength\abovedisplayskip{2pt}
\setlength\belowdisplayskip{2pt}
\begin{equation}
\mathcal{L}_{\mathbf{\hat{C}}}^{B}=\sum_{\mathbf{I}, \mathbf{T} \in \mathcal{D}}\sum_{i=1}^N \theta^{N-i}\mathcal{L}_{BCE}(\mathbf{\hat{C}}_i, \mathbf{C}_{gt}),
\label{eqn:loss_cmap_bce}
\end{equation}
\dz{where $\theta$ is set to 0.85. 
We follow the method in~\cite{qian2018attentive} to obtain the GT RCMaps: $\mathbf{C}_{gt}$. First, we subtract the input image $\mathbf{I}$ with its corresponding transmission image $\mathbf{T}$ to obtain an absolute residual image $\mathbf{\Bar{R}}= \|\mathbf{I} - \mathbf{T}\|$, and then apply a threshold $\gamma$ to $\mathbf{\Bar{R}}^ {gray}$ to get a binary mask as $\mathbf{C}_{gt}$, where $\gamma$ is set to $\max\{\mathbf{\Bar{R}}^{gray}_{\min} + \beta * (\mathbf{\Bar{R}}^{gray}_{\max} - \mathbf{\Bar{R}}^{gray}_{\min}), \beta\}$, and $\mathbf{\Bar{R}}^{gray}$ is the gray-scale  $\mathbf{\Bar{R}}$. We re-train our model using $\mathcal{L}_{\mathbf{\hat{C}}}^B$ and set the parameter $\beta$ to 0.3 for the best performance. The choice of $\beta$ is evaluated in Sec.4 of the supplemental material.}}

\renewcommand\arraystretch{1}
\begin{table}[t]
\begin{center}
\resizebox{0.95\columnwidth}{!}{
\begin{tabular}{ccc|cc|cc}
\toprule[1.1pt]
\multirow{2}*{Model} & \multicolumn{2}{c}{\textit{SIR}$^2$~\cite{2017Benchmarking}} & \multicolumn{2}{c}{Zhang \etal.~\cite{zhang2018single}} & \multicolumn{2}{c}{Li \etal.~\cite{li2020single}}\\

\cmidrule[0.5pt]{2-7}
& PSNR & SSIM & PSNR & SSIM & PSNR & SSIM\\
\hline

w/o $\mathcal{L}_{pixel}$ & 23.365 & 0.888 & 22.113 & 0.788 & 21.587 & 0.788\\

w/o $\mathcal{L}_{SSIM}$ & 15.653 & 0.673 & 17.652 & 0.746 & 15.197 & 0.642\\

w/o $\mathcal{L}_{P}$ & 22.725 & 0.880 & 22.219 & 0.804 & 22.778 & \textbf{0.812}\\

w/o $\mathcal{L}_{comp}$ & 23.774 & 0.894 & 22.453 & 0.807 & 23.036 & 0.801\\

w/o $\mathcal{L}_{adv}$ & 23.571 & 0.889 & 22.480 & 0.803 & 23.240 & 0.800\\


$\mathcal{L}_{\mathbf{\hat{C}}} \rightarrow \mathcal{L}^B_{\mathbf{\hat{C}}}$ & 24.106 & 0.894 & 21.296 & 0.776 & 22.176 & 0.788  \\



$\mathcal{L}^B_{\mathbf{\hat{C}}}$ \& RKI & 24.114 & 0.899 & 21.123 & 0.778 & 22.253 & 0.788 \\

\hline

Ours & \textbf{24.117} & \textbf{0.901} & \textbf{23.338} & \textbf{0.812} & \textbf{23.451} & 0.808\\


\bottomrule[1.5pt]
\end{tabular}}  
\end{center}
\vspace{-5pt}
\caption{Ablation study on Loss terms. w/o $\mathcal{L}$: we remove each loss term $L$ and evaluate the corresponding re-trained model to check its influence on the reflection removal results. $\mathcal{L}_{\mathbf{\hat{C}}} \rightarrow \mathcal{L}^B_{\mathbf{\hat{C}}}$: replace $\mathcal{L}_{\mathbf{\hat{C}}}$ with $\mathcal{L}^B_{\mathbf{\hat{C}}}$.   $\mathcal{L}^{B}_{\hat{C}}$ \& RKI: replace $\mathcal{L}_{\hat{C}}$, LKI with $\mathcal{L}^{B}_{\hat{C}}$ and RKI respectively.}
\label{tab:loss_ablation}
\vspace{-18pt}
\end{table}


\dz{
As shown in Fig.~\ref{fig:calculate_rcmap}, such a simple thresholding method occasionally generates GT RCMaps (the third column in Fig.~\ref{fig:calculate_rcmap}) that incorrectly label some transmission-dominated regions with the reflection-dominated regions, even with relative intensity method (the last column in Fig.~\ref{fig:calculate_rcmap}, $\beta=0.1$). We hypothesize that it is why our model based on $\mathcal{L}_{\hat{C}}$ surpasses the other two variants that use Eq.~\ref{eqn:loss_cmap_bce} and RKI in PSNR/SSIM scores, as shown in the last three rows of Tab.~\ref{tab:loss_ablation}. Besides, the 
third column of Fig.~\ref{fig:compar_rcmap} illustrates that the RCMap obtained under the supervision of $\mathcal{L}_{\mathbf{\hat{C}}}^{B}$ ($\beta$ = 0.3) contains mislabeling errors (red boxes), resulting in damages to non-reflection regions when estimating $\mathbf{\hat{T}}$. In Sec.5 of the supplemental material, we evaluate the detection accuracy further and show that our method can well protect non-reflection regions. }

\vspace{-5pt}
\section{Conclusion}
We develop a location-aware SIRR network to improve the quality of SIRR results substantially. Its key feature is that we leverage learned Laplacian features that can emphasize the strong reflections' boundaries to locate and remove strong reflections, such as reflected highlights. The network has an RDM that takes multi-scale Laplacian features as inputs to detect reflections roughly. In the future, we plan to simplify our network's design further to save the number of parameters and improve its inference speed on the mobile computing platform.

\vspace{5pt}
\noindent\textbf{Acknowledgements}:\quad We thank the reviewers for their constructive comments. Weiwei Xu is supported by NSFC (No.~61732016). This work was partially supported by a GRF grant from RGC of
Hong Kong (Ref.: 11205620).

{\small
\bibliographystyle{ieee_fullname}
\bibliography{egbib}
}

\end{document}